\documentclass{article}

\usepackage[preprint]{corl_2022} 
\usepackage{graphicx}
\usepackage{graphicx}
\usepackage{bm}
\usepackage{booktabs}
\usepackage{times}
\usepackage{epsfig}
\usepackage[ruled,vlined]{algorithm2e}
\usepackage{ctable}
\usepackage{caption}
\usepackage{subcaption}

\usepackage{comment}
\usepackage{amsmath,amssymb} 
\usepackage{color}
\usepackage{multirow}
\usepackage{multicol}
\usepackage[autostyle]{csquotes} 
\usepackage[capitalize]{cleveref}
\usepackage{dsfont}
\newcommand{\xhdr}[1]{ \noindent {\textbf{#1}}}
\newsavebox{\measurebox}

\newcolumntype{L}[1]{>{\raggedright\let\newline\\\arraybackslash\hspace{0pt}}m{#1}}
\newcolumntype{C}[1]{>{\centering\let\newline\\\arraybackslash\hspace{0pt}}m{#1}}
\newcolumntype{R}[1]{>{\raggedleft\let\newline\\\arraybackslash\hspace{0pt}}m{#1}}

\newcommand{\beginsupplement}{%
        \setcounter{table}{0}
        \renewcommand{\thetable}{S\arabic{table}}%
        \setcounter{figure}{0}
        \renewcommand{\thefigure}{S\arabic{figure}}%
        \setcounter{section}{0}
     }

\title{Learning Road Scene-level Representations via Semantic Region Prediction}

\author{
  Zihao Xiao, Alan Yuille\\
  Department of Computer Science\\
  Johns Hopkins University\\
  \texttt{\{zxiao10,ayuille1\}@jhu.edu}
  \And
  Yi-Ting Chen \\
  Department of Computer Science\\
  National Yang Ming Chiao Tung University \\
  \texttt{ychen@cs.nycu.edu.tw}
}

\begin{document}
\maketitle


\begin{abstract}
%
%
%
%
%
In this work, we tackle two vital tasks in automated driving systems, i.e., driver intent prediction and risk object identification from egocentric images.
Mainly, we investigate the question: what would be good road scene-level representations for these two tasks?
We contend that a scene-level representation must capture higher-level semantic and geometric representations of traffic scenes around ego-vehicle while performing actions to their destinations.
To this end, we introduce the representation of semantic regions, which are areas where ego-vehicles visit while taking an afforded action (e.g., left-turn at 4-way intersections). 
%
%
%
%
%
%
We propose to learn scene-level representations via a novel semantic region prediction task and an automatic semantic region labeling algorithm.
%
%
Extensive evaluations are conducted on the HDD and nuScenes datasets, and the learned representations lead to state-of-the-art performance for driver intention prediction and risk object identification. 
%
%
\end{abstract}



\section{Introduction}
%
%
\begin{figure}[t]
\centering
\includegraphics[width = \columnwidth,trim={0 0 0 3.5cm},clip]{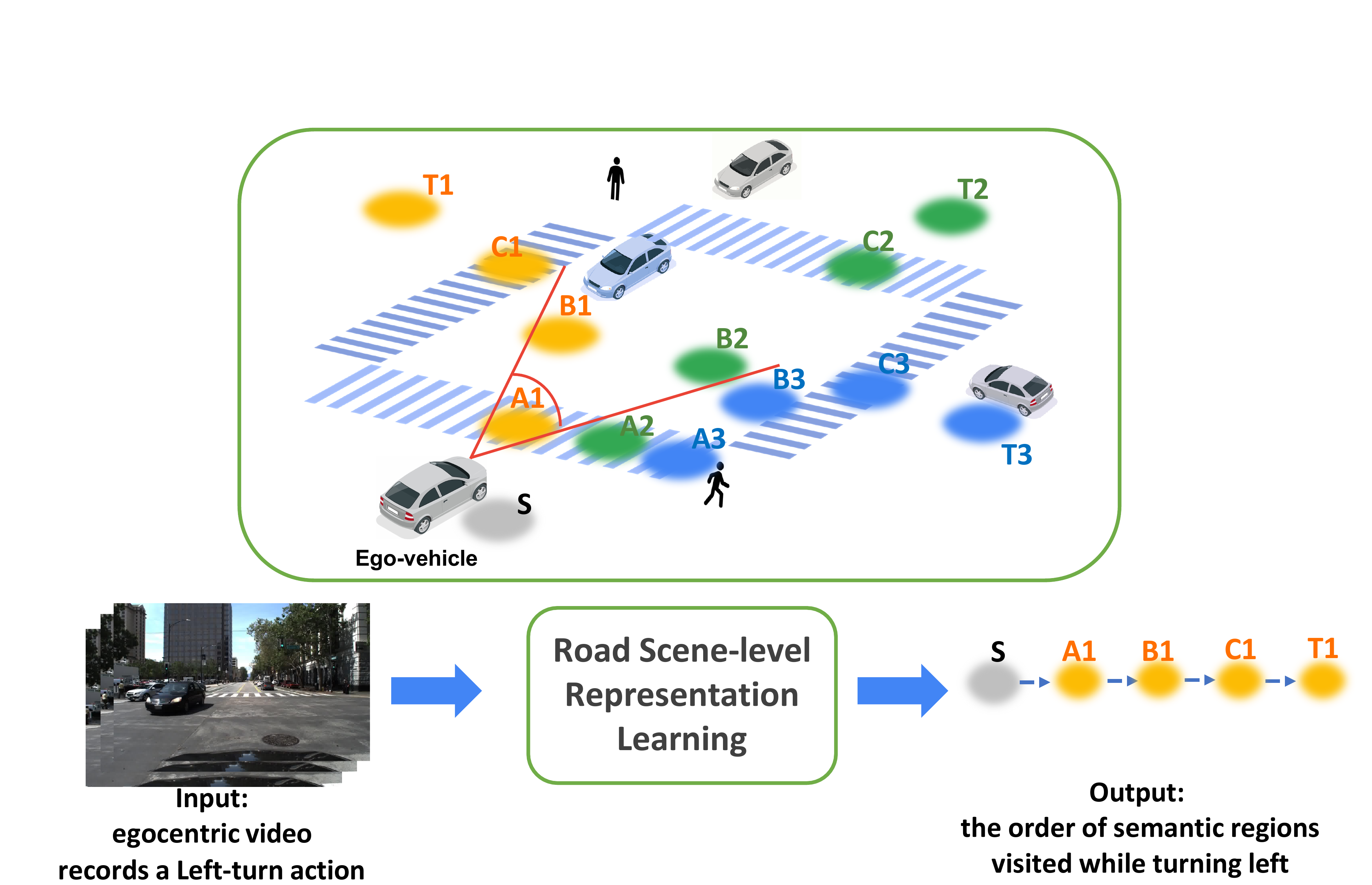} \caption{\textbf{Main idea.} 
We contend that a scene-level representation must capture higher-level semantic and geometric representations of traffic scenes around ego-vehicles while performing actions to their destinations in order to reason about the larger scenes.
We propose semantic region, a novel representation that represents areas where ego vehicles visit while taking an afforded action.
We associate egocentric images, representing views from different locations of road scenes under specific actions, with the corresponding semantic regions.
We cast road scene-level representation learning as semantic region prediction and demonstrate the learned representations are effective for driver intention prediction and risk object identification.
}
\label{fig:main}
\end{figure}


For automated driving systems (e.g., advanced driver assist systems, ADAS) to navigate highly interactive scenarios, they must be able to perceive states of traffic elements, forecast traffic situations, identify potential hazards, and plan the corresponding actions. 
%
%
The field has made substantial progress
in the past few years~\cite{Zhang2013ICCV,Geiger_Scene_PAMI2014,cityscapes,Zhou_cvpr2017,mappilary,HeCVPR2017,ETDR2020,yolov4,Wojke2017simple,Schulter_ECCV2018,Wang_top_CVPR2019,Yang_VIS2019,TraPHic_CVPR2019,Roddick_PON_CVPR2020,Philion_lift_eccv2020,Guizilini,Axial-DeepLab,Jung_iccv2021,malawade2021roadscene2vec,Neumann_CVPR2021}.
%
%
In this work, we focus on improving the performance of driver intention prediction~\cite{Doshi_driver_intent_2011,Jain_driver_maneuver_iccv2015,Jain_structuralRNN_CVPR2016} and risk object identification~\cite{Ohn-Bar_importance_2017,Gao_goal_icra2019,wang_object_icra2019,li2020make,semi-roi} from egocentric videos.
Solving both tasks from egocentric videos is crucial for safety systems such as ADAS, where front-facing cameras are the primary device.

Existing works for both tasks~\cite{Jain_driver_maneuver_iccv2015,Ohn-Bar_importance_2017,wang_object_icra2019,li2020make,semi-roi}  utilize image annotations of the tasks (intent prediction and potential hazard identification) and object cues from object detection to train networks in a supervised learning manner. 
Additionally, the authors of~\cite{Jain_driver_maneuver_iccv2015,9294181} leverage temporal models such as LSTM~\cite{hochreiter1997long} and ConvLSTM~\cite{convlstm}, and spatial-temporal interaction between traffic participants are modeled using spatial-temporal graph~\cite{Jain_structuralRNN_CVPR2016} and graph convolutional networks~\cite{Kipf_GCN_iclr2017} to further improve the performance of tasks.
While promising results are demonstrated, the learned representations are ineffective to the trained task. 
Moreover, the representations only encode road scenes in the nearby locality. 
We contend that a scene-level representation must capture higher-level semantic and geometric representations of traffic scenes around ego-vehicles while performing actions to their destinations in order to reason about the larger scenes.
We introduce a novel representation called the semantic region, as shown in \cref{fig:main}. 
Semantic regions are areas where ego-vehicles visit while taking an afforded action to their destination. The birth of semantic region is motivated by road affordance (i.e., possible actions that a vehicle can take in an environment).
For instance, while turning left at an intersection, the vehicle visits the semantic regions (in yellow), i.e., the \textit{crosswalk} near the ego vehicle, \textit{the area of intersection}, and the \textit{crosswalk} on the left sequentially.
If different afforded actions are taken, different semantic regions will be visited. 
Note that different road topologies (e.g., 3-way intersections and straight roads) afford different actions. Our insight is that there are finite regions that a vehicle visits when taking action afforded the underlying road topology. 
Therefore, we associate egocentric images, representing views from different locations of road scenes under certain actions, to the corresponding semantic regions.

\begin{figure}
\centering
\captionsetup[subfigure]{justification=centering}
\sbox{\measurebox}{%
  \begin{minipage}[b]{.4\textwidth}
  \subfloat
    [4-way Intersection and associated egocentric view images][Intersection: 4-way]
    {\label{fig:SR_4_way}
    \includegraphics[width=\textwidth,,height=4cm]{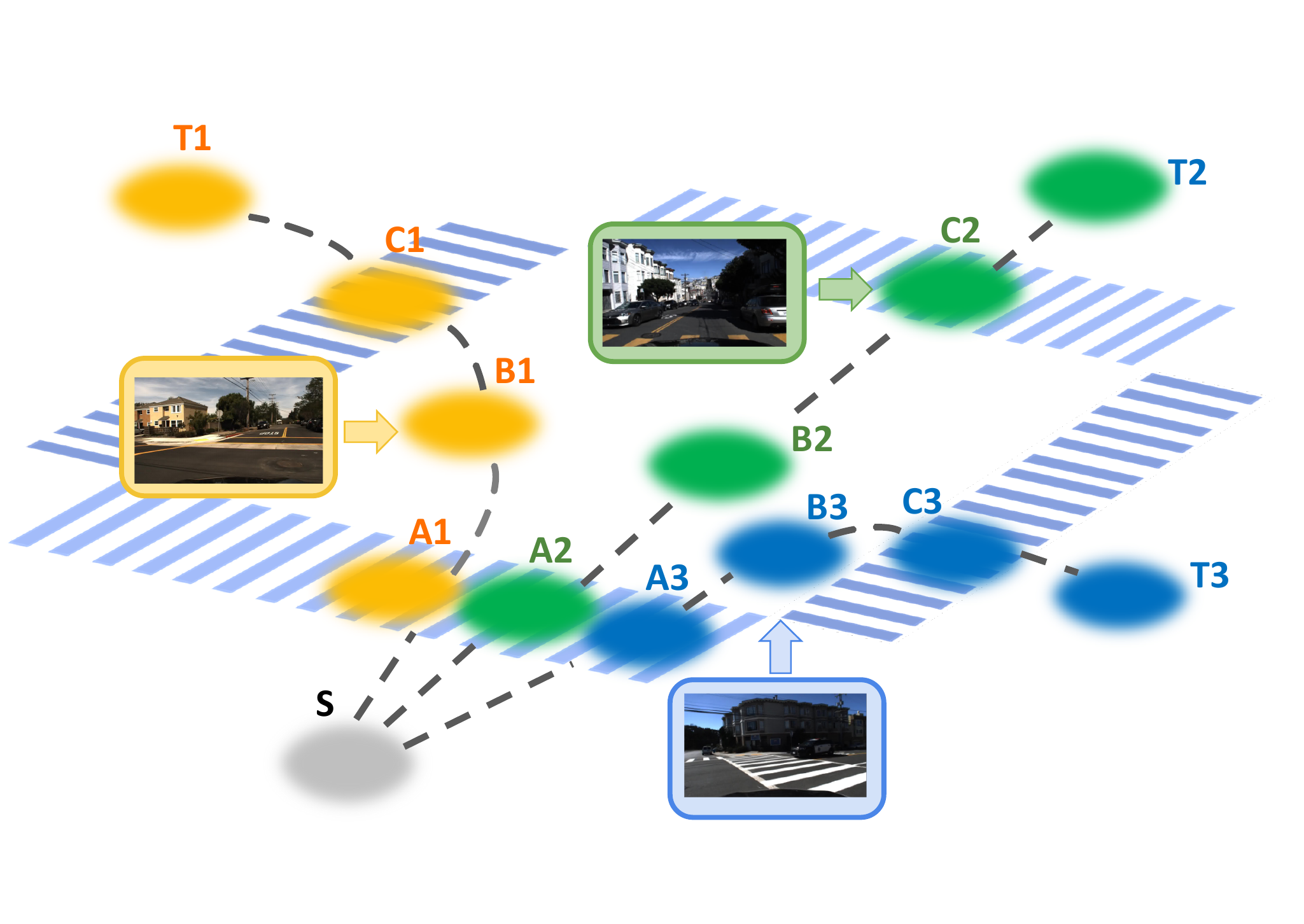}
    }
\vfill
  \subfloat
    [4-way Intersection and associated egocentric view images][Non-intersection: Lane Change]
    {\label{fig:SR_lane_change}
    \includegraphics[width=\textwidth,,height=3cm]{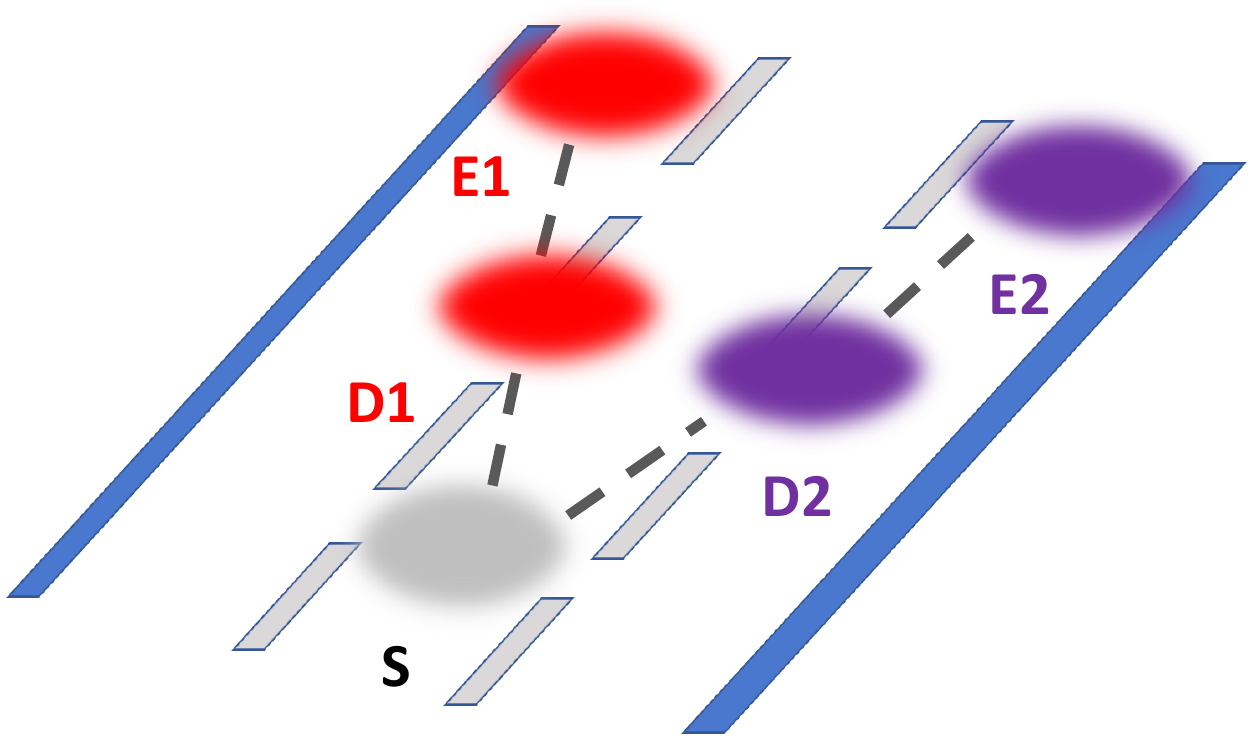}
    }
  \end{minipage}}
\usebox{\measurebox}
\qquad
\begin{minipage}[b][\ht\measurebox][s]{.35\textwidth}
\centering
\subfloat[Left-turn/Straight Intersection][Intersection: Left-turn/Straight]
  {\label{fig:SR_LS}\includegraphics[width=\textwidth,height=4cm]{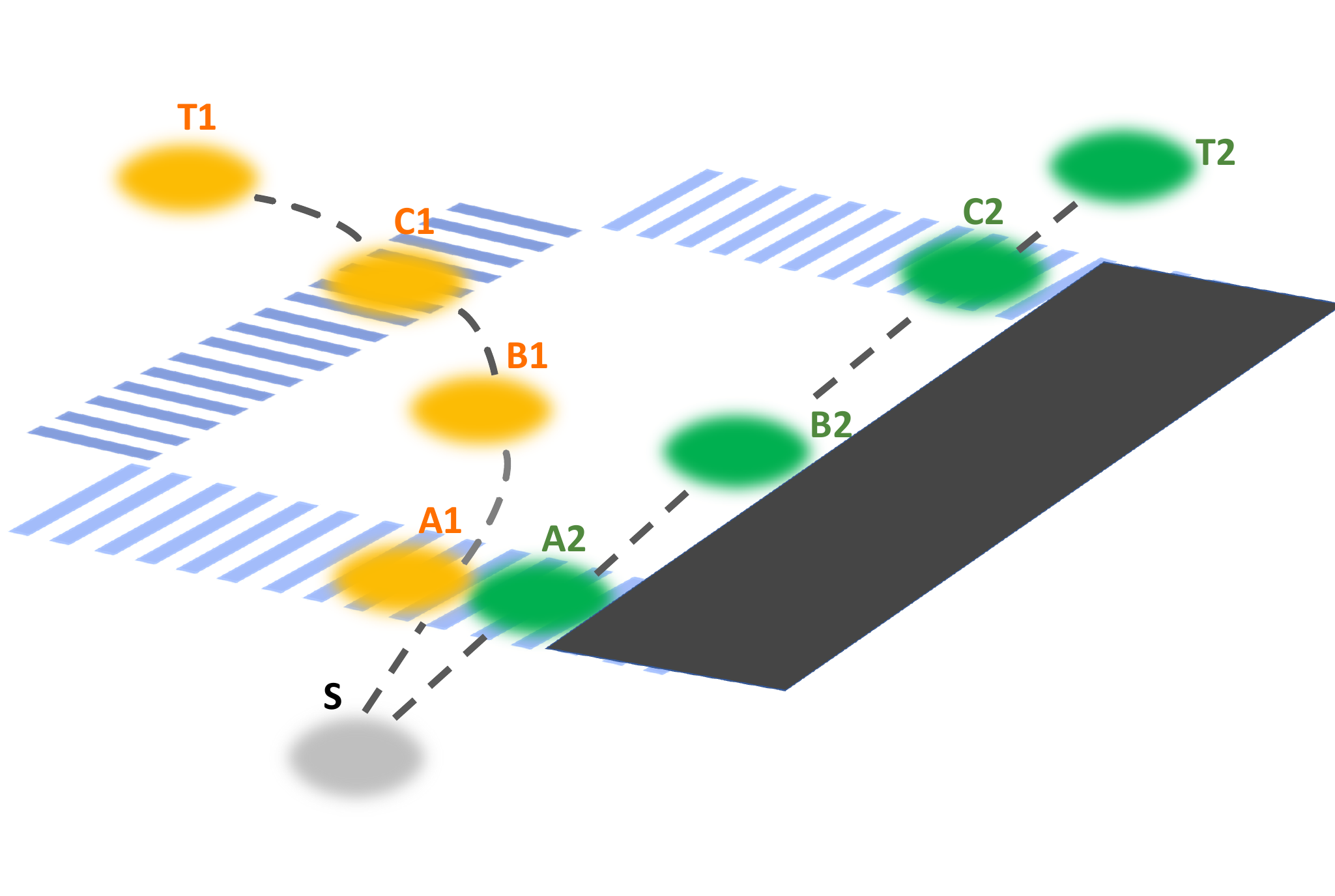}}
\vfill
\subfloat
  [Intersection: Left-turn/Right-turn]
  {\label{fig:SR_LR}\includegraphics[width=\textwidth,height=3cm]{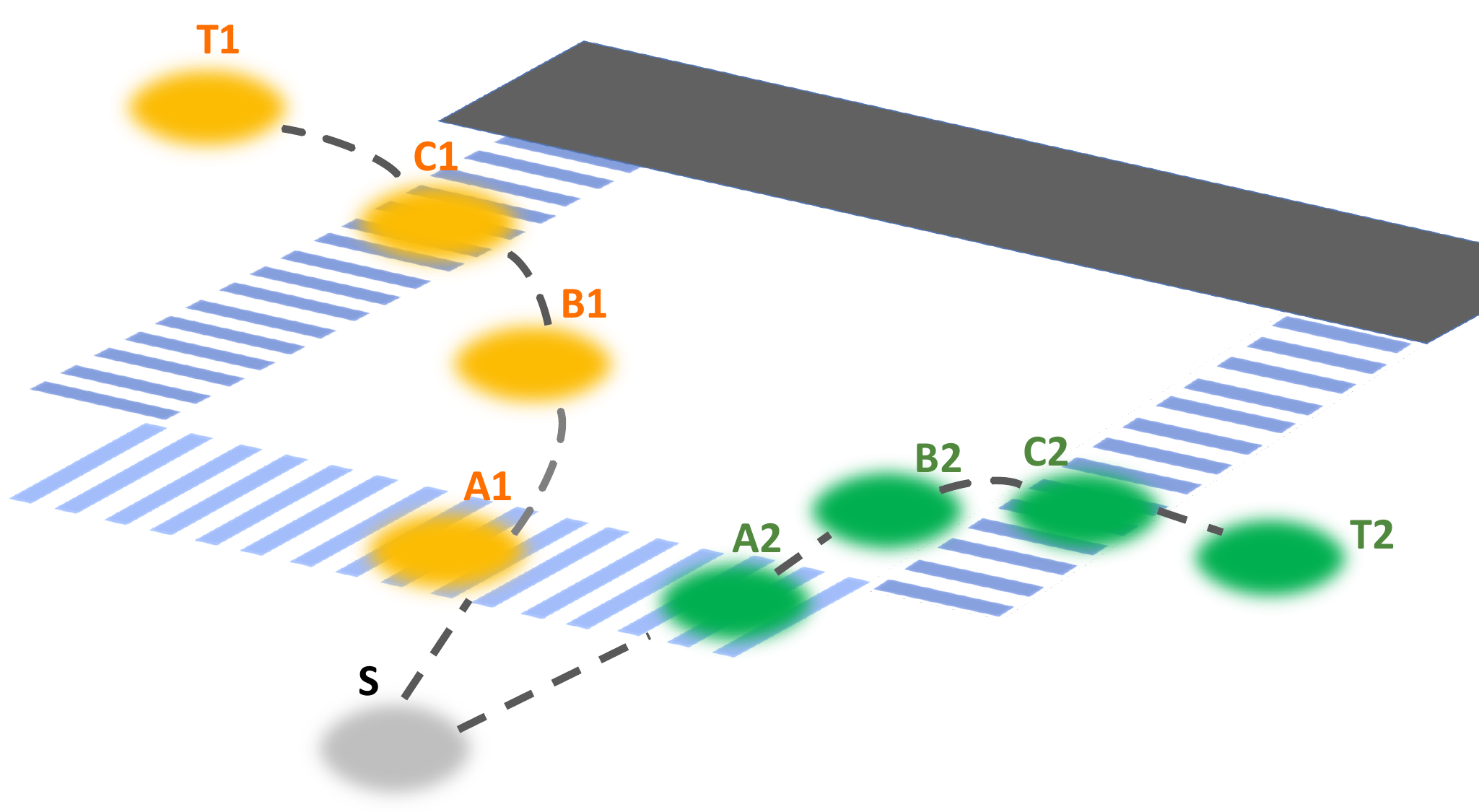}}
\end{minipage}
\caption{\textbf{Semantic regions in different types of road topology.} Semantic regions are areas where  ego-vehicles visit while taking actions afforded the underlying road topology. For instance, at 4-way intersections, three actions (i.e., Left-turn,  Straight,  Right-turn) are afforded. Given egocentric images while performing an afforded action, we associate them with the corresponding semantic regions.
In this work, we cover a wide range of road topologies, i.e., 4-way/3-way intersections and straight roads with multiple lanes.
%
%
}
\label{fig:diff_road_topology}
\end{figure}

We cast scene-level representation learning as semantic region prediction.
Specifically, the model predicts future semantic regions sequentially, given historical observations before turning left.
For instance, as shown in \cref{fig:SR_4_way}, given egocentric images representing semantic regions $S$ and $A_{1}$, the task aims to predict future semantic regions $B_{1}$, $C_{1}$, and $T_{1}$ in sequential order.
%
%
To enable representation learning, we design an automatic semantic region annotation strategy to label every egocentric image collected in intersections with the corresponding semantic region, which reduces the annotation burden.
%
%
%

%
%
%
%
%

We demonstrate the effectiveness of the scene-level representation learning framework on driver intention prediction~\cite{Jain_driver_maneuver_iccv2015} and risk object identification~\cite{li2020make}.
%
We achieve superior performance compared to strong baselines for driver intention prediction on the HDD dataset~\cite{Ramanishka_behavior_CVPR_2018}.  
Furthermore, we show favorable generalization capability without additional training on nuScenes~\cite{caesar2020nuscenes}.
%
Moreover, our framework obtains state-of-the-art performance for risk object identification.
Specifically, we boost the current best-performing algorithm~\cite{li2020make} by 6\%.
%
%

Our contributions are summarized as follows. First, we propose a novel representation called semantic region, which aims to capture higher-level semantic and geometric representations of traffic scenes around ego vehicles while performing actions to their destination. Second, we cast scene-level representation learning as semantic region prediction (SRP) and propose an automatic labeling algorithm for intersections to reduce annotation burdens. Third, we conduct extensive evaluations on the HDD and nuScenes datasets to prove that the effectiveness of the learned representations leads to significant improvements in driver intention prediction and risk object identification. 

\section{Related Work}

\xhdr{Driver Intention Prediction.} 
Advanced driver-assistance systems predict driver intention~\cite{Doshi_driver_intent_2011,Jain_driver_maneuver_iccv2015,Phillips_generalizable_IV2017,Zyner_RNN_intent_RAL2018,Casas_IntentNet_corl2018,Hu_semantic_IV2018} to avoid potential hazards.
Doshi et al.,~\cite{Doshi_driver_intent_2011} predict driver's intent via reasoning distances to lane markings and vehicle dynamics for driver intention prediction in highway scenarios.
%
In the Brain4Car project~\cite{Jain_driver_maneuver_iccv2015,Jain2016RecurrentNN,Jain_structuralRNN_CVPR2016}, multi-sensory signals, including GPS and street maps, are used for anticipation of driving maneuvers. 
Similarly, pre-computed road topology maps around intersections are utilized to extract features such as ego position and
dynamics, distance to surrounding traffic participants, and legal actions at the upcoming intersection to predict driver intention~\cite{Phillips_generalizable_IV2017}.
Recently, Casas et al.,~\cite{Casas_IntentNet_corl2018} leverage rasterized HD maps as input deep neural networks for intent prediction.
Instead of formulating intent prediction as a recognition problem, Hu et al.,~\cite{Hu_semantic_IV2018} formulate intention prediction as entering an \textit{insertion area} defined on a pre-computed road topology map. 
For instance, if the intent is turning left, the corresponding \textit{insertion area} is $T_1$ as shown in \cref{fig:SR_4_way}.
%
%
%
Unlike existing methods exploiting pre-computed road topology, we learn road scene-level representations via semantic region prediction, which captures higher-level semantic and geometric representations of traffic scenes around ego vehicles while performing actions to their destinations.
We empirically demonstrate the value of learned representations.
%



\xhdr{Risk Object Identification.}
%
%
The goal of risk object identification is to identify object(s) that impact ego-vehicle navigation~\cite{Ohn-Bar_importance_2017,kim2017interpretable,Palazzi_dreye_pami2018,Gao_goal_icra2019,wang_object_icra2019,li_gcn_icra2020,Zhang_GCN_icra2020,li2020make}.
The authors of~\cite{Ohn-Bar_importance_2017,Gao_goal_icra2019,Zhang_GCN_icra2020} construct datasets with object importance annotation, and supervised learning-based algorithms are designed and trained to identify risk/important objects.
The task can be formulated as selecting regions/objects  with high activations in visual attention heat maps learned from end-to-end driving models~\cite{kim2017interpretable,wang_object_icra2019,li_gcn_icra2020}.
Recently, Li et al.,~\cite{li2020make} formulate risk object identification as a cause-effect problem~\cite{Pearl_causality_2009}.
They propose a two-stage risk object identification framework and demonstrate favorable performance over~\cite{kim2017interpretable,wang_object_icra2019}.
In this work, we extend~\cite{li2020make} with the learned road scene-level representations because driver intention is crucial for risk/important object identification~\cite{Gao_goal_icra2019}.
%
%
Note that they~\cite{Gao_goal_icra2019} assume that the planned path is given. In this work, we tackle a more challenging setting where driver intention is unknown and should be inferred from egocentric images. 

\section{Automatic Semantic Region Labeling}
\begin{figure}[t]
\centering
\includegraphics[width=\textwidth]{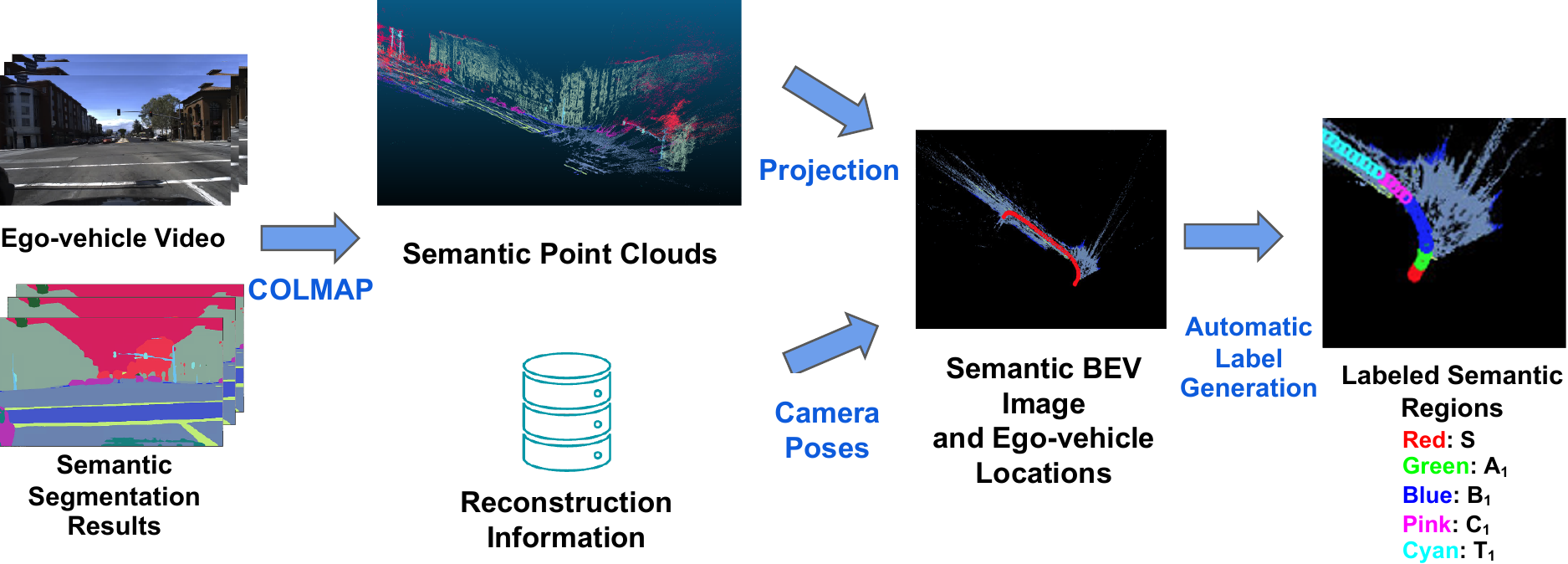}
    \caption{\textbf{Automatic labeling of semantic regions in intersections.} We show an example of generating labels of semantic regions from a \textit{Left-turn} egocentric video sequence. The semantic regions are consistent with the ones in \cref{fig:SR_4_way}. Best viewed in color.}
    \label{fig:COLMAP}
\end{figure}

We propose an automatic labeling strategy to ease the burdens.
The overall generation process is depicted in \cref{fig:COLMAP}.
Specifically, a three-step strategy is proposed. 
First, given egocentric videos collected while taking afforded actions (i.e., \textit{Left-turn}, \textit{Straight}, \textit{Right-turn}, and \textit{Lane-change}) without interacting with traffic participants from the HDD dataset~\cite{Ramanishka_behavior_CVPR_2018}\footnote{The HDD dataset provides large-scale annotations of afforded actions.}, we apply COLMAP~\cite{schoenberger2016sfm}, to obtain a dense 3D reconstruction and camera poses. 
%
In addition, semantic segmentation~\cite{MVD2017} is applied to every egocentric image.
Second, each 3D point is projected onto images so that the point is visible to obtain the corresponding semantic candidates. 
Then, a simple winner-take-all strategy is used to determine the final label.
%
We project the semantic 3D point cloud to the ground plane to obtain a semantic Bird-Eye-View (BEV) image. 
Third, we label the semantic region of each camera pose with the information from semantic BEV image. For example, in intersections, we assume that ego vehicles will visit two \textit{crosswalks} sequentially while taking afforded actions.
Camera poses that overlap with the first \textit{crosswalk} and the second \textit{crosswalk} are denoted as $A_{i}$ and $C_{i}$, respectively.
The poses located between $A_{i}$ and $C_{i}$ are $B_{i}$.
Camera poses located before the first \textit{crosswalk} and the second \textit{crosswalk} as $S_{i}$ and $T_{i}$, respectively. Each index $i$ represents an afforded action.
Last but not least, while the results of COLMAP and semantic segmentation are generally well, we use two additional criteria to select good samples: 1) 3D reconstruction is successful, and 2) reconstructed camera poses form a coherent trajectory. 
Note that the algorithm is in general applicable for different topologies.
However, we observed failures for \textit{lane-change} in non-intersection due to inaccurate 3D reconstruction.
Therefore, we manually annotate videos that ego-vehicles perfrom lane-change. Details of automatic semantic region labeling are provided in the supplementary
materials.

\label{sec:ASL}
\section{Methodology}
In this section, we discuss the details of road scene-level representation learning from egocentric video via semantic region prediction. In addition, we illustrate how to transfer the learned representation to two downstream tasks, i.e., driver intention prediction and risk object identification.
\subsection{Scene-level Representation Learning via Semantic Region Prediction}
%
We contend that a scene-level representation must capture higher-level semantic and geometric representations of traffic scenes around ego-vehicle while performing actions to their destinations.
%
Thus, we proposed the representation called semantic region, which is a high-level abstraction of road affordance.  
%
%
We expect a model capturing the association between the temporal evolution of egocentric views and semantic regions. 
To this end, We cast the egocentric road scene affordance representation learning as a semantic region prediction task. We build our Semantic Region Prediction (SRP) cell based on TRN cells ~\cite{XuOnlineICCV2019}. 
The STA (spatio-temporal accumulator) in the TRN cell makes use of predicted future cues from the temporal decoder and the accumulated historical information to form better action representations. To have a unified framework, We make the following changes to TRN cells. First, we replace the action classifier with two semantic region classifiers for both intersections and non-intersections. Second, in the decoder, the predicted logits of two semantic region classifiers are fused into the input of the next time frame after increasing dimensions with fully connected(FC) layers. Third, we add a topology classifier to determine whether the ego-vehicle is at intersections (4-way, 3-way) or non-intersections (straight road or curve). Note that our design is similar to CILRS~\cite{codevilla2018end}, a command-conditional imitation learning framework. In our case, we select the corresponding set of semantic regions (as shown in \cref{fig:diff_road_topology}) based on the prediction of topology type.

\begin{figure*}[t]
\centering
\includegraphics[width=\textwidth]{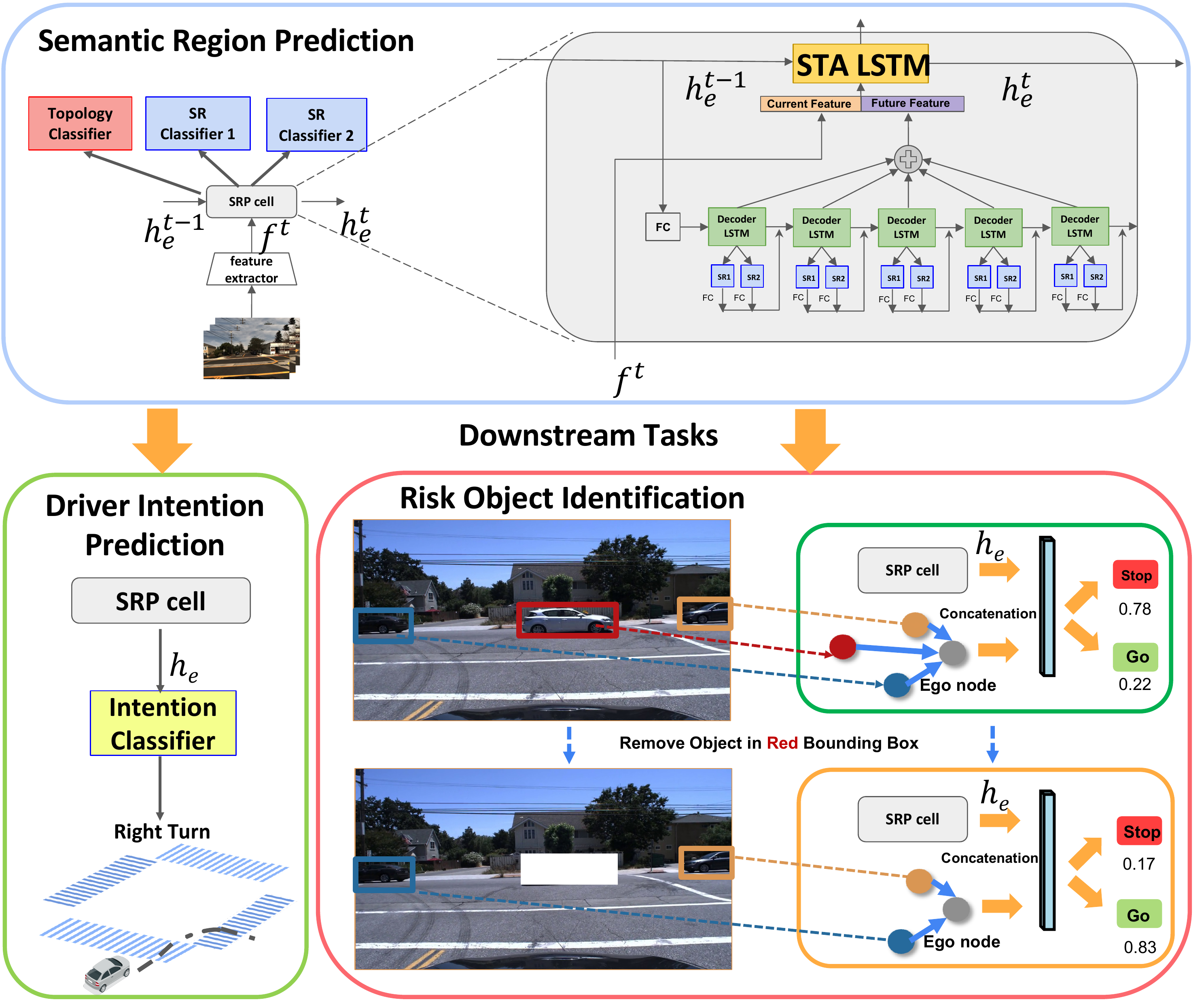}

\caption{\textbf{The proposed network architecture for semantic region prediction and models for downstream tasks.} We propose to learn road scene representation via Semantic Region Prediction (SRP). The hidden state of the SRP cell serves as road scene representation and is utilized in two downstream tasks: driver intention prediction and risk object identification. 
}
\label{fig:Architecture}
\end{figure*}

With SRP cells, our network takes $t_e$ historical frames as input. For each frame, topology type (i.e., whether it is in an intersection), the current semantic region as well as  $t_d$ future semantic regions are predicted. We have separate semantic region classifiers for intersections and non-intersections. During training, we only compute losses for the one that matches the ground truth topology type. 
The semantic region prediction loss $\mathcal{L}$ is defined as 
\begin{equation}
    \mathcal{L}=\sum_{t}^{t_e}\mathbf{l}(z_{t},o_{t}) +  \sum_{i=0}^{1}\mathds{1}_{o_{t}=i}\Bigl(\sum_{t}^{t_e}\mathbf{l}(y_{t}^{i,e},s_{t}^{i}) + \frac{1}{{t_d}}\sum_{m=0}^{t_d}\mathbf{l}(y_{t}^{i,d},s_{t+m}^{i})\Bigl)
\label{eq:lr_sr}
\end{equation}
where $i$ $\in{\{0,1\}}$  denotes the $i$th semantic region classifier, $z_t$ denotes the topology prediction, and  $y_{t}^{i,e}$ and $y_{t}^{i,d}$ are the semantic region prediction based on the hidden state of STA and decoder respectively for topology classifier $i$. $\mathbf{l}$ is the Cross-entropy loss, $\mathds{1}$ is the indicator function, $o_t$ is the ground truth topology type, and $s_t^{i}$ is the ground truth of semantic regions derived from \cref{sec:ASL}.  The overall architecture is depicted in \cref{fig:Architecture}. In practice, it is not necessary to observe all semantic regions in one video clip. For instance, in an intersection, when there are no crosswalks, $A_i$ and $C_i$ will not be presented. For a left-turn vehicle at intersections without a crosswalk, the semantic region sequence  will be $S- B_1-T_1$.
The hidden state $\mathbf{h}_e^t$ of STA contains rich information about the road scene.
Next, we will show how to incorporate the learned representation into downstream tasks.

%
%
 
%

%

\subsection{SRP-guided Driver Intention Prediction}

We follow the definition of anticipation in \cite{Jain2016RecurrentNN} to define driver intention prediction.
Formally, given an sequence of egocentric observations$\{\mathbf{x}^1,\mathbf{x}^2,...,\mathbf{x}^t\}$, our goal is to predict the future intention $\mathbf{y}_{int}^T$, where $T>t$. Driver intention prediction  benefits downstream applications like risk assessment \cite{lefevre2014survey}. 
There are 5 different types of intentions in our setting (i.e., Left-turn, Straight, Right-turn, Left-lane-change, Right-lane-change).
We add an intention classifier on top of the hidden state of the STA, $\mathbf{h}_e^t$ in SRP,
\begin{equation}
    \mathbf{y}_{int}^T = \text{softmax} (\mathbf{W}_{int}^\top \mathbf{h}_e^t+ \mathbf{b}_{int})
\end{equation}
where $\boldsymbol{\mathbf{W}_{int}^\top}$ and $\mathbf{b}_{int}$ are the weight and bias terms in the intention classifier, respectively. We name the driver intention prediction model SRP-INT. 


\subsection{SRP-guided Risk Object Identification}
\label{subsec: t_risk_object}
The risk object identification task was first introduced in \cite{li2020make}. A Risk object is defined as the one influencing the behavior of the ego-vehicle most in each frame. Given an egocentric video$\{\mathbf{x}^1,\mathbf{x}^2,...,\mathbf{x}^t\}$, the goal of risk object identification is to output$\{\mathbf{b}^1,\mathbf{b}^2,...,\mathbf{b}^t\}$, where $\mathbf{b}^j$, $ j\in[1,t]$ is the bounding box of the risk object in the $j$-th frame.  The authors of \cite{li2020make} proposed a two-stage framework to solve the problem. In the first stage, they trained an object-level manipulable model to predict the driver behavior by incorporating partial CNNs~\cite{liu2018image}. In the second stage, they iterated through the risk object candidate list and intervened in the input video to simulate scenarios without the presence of a candidate. The simulated scenarios were passed into the driver behavior model. The object causing the maximum driving behavior change was their risk object prediction. The ego-representation in \cite{li2020make} takes a very important role because it captures the information from the image frame and the messages from all the objects. The representation in time $t$, i.e., the last time step, can be written as
\begin{equation}
    \mathbf{g}_{e}^t = \mathbf{g}_{f}^t \oplus \frac{1}{N} \sum_{k=1}^N \mathbf{g}_{k}^t
\end{equation}
where $\mathbf{g}_{f}^t$ is the representation of the image frame, $\mathbf{g}_{k}^t, k \in[1,N]$ in the representation for each object, $\oplus$ indicates a concatenation operation, and $\mathbf{g}_{e}^t$ is the final ego-representation in \cite{li2020make}. 

We propose SRP-ROI by fusing SRP with the model in \cite{li2020make}. {We argue that road scene-level information can benefit the risk object identification task, and propose an SRP-guided representation:}
\begin{equation}
    \mathbf{g}_{e}^t = \big((\mathbf{W}_{ego}\mathbf{g}_{f}^t \oplus \frac{1}{N} \sum_{k=1}^N \mathbf{g}_{k}^t)+\mathbf{b}_{ego}^t\big) \oplus \boldsymbol{h}_{e}^t
    \label{con:ROI}
\end{equation}
where $\mathbf{W}_{ego}$ and $\mathbf{b}_{ego}$ are the weights and bias terms of a fully connected layer respectively. We follow the two-stage framework in \cite{li2020make} and evaluate our SRP-ROI model on two challenging dynamic risk object categories: crossing vehicles and crossing pedestrians.


\section{Experiments}
\subsection{Semantic Region Prediction}

\xhdr{Data Collection and Annotation.} We collect video clips of Left-turn, Straight, Right-turn, Left-lane-change, and Right-lane-change from the HDD dataset to train our semantic region predictor. For each video clip, we manually label the topology type. Labels of semantic regions at the intersections are automatically generated with the methods proposed in \cref{sec:ASL}. The semantic regions for non-intersections are annotated by humans. For each video clip, we apply a sliding-window method to obtain training samples. For each sample, we have annotations including topology type, current, and future semantic region labels.

\xhdr{Implementation Details and Results.} We leverage ResNet50 [23] pre-trained on Mapillary Vistas [45] dataset as the feature extractor. Our SRP takes $l_e$ = 3 historical frames
as input. For each frame, $l_d$ = 5 future semantic regions, as well as topology type, are predicted. As shown in \cref{fig:SR_4_way} and \cref{fig:SR_lane_change}, the number of semantic regions in intersection and non-intersections are 13 and 5, respectively. We use Adam optimizer \cite{adam2015} with default parameters, a learning rate of $0.0001$, and weight decay of $0.0005$. The model is trained for 60 epochs. We train the model with the loss function in \cref{eq:lr_sr}. The performances are shown in \cref{table:SR}. Macro Average Precision, Micro Average Precision, and mAP are chosen as the evaluation metrics.

\begin{table}[]
\small
\centering
\begin{tabular}{c|c|c|c|c}
\toprule[2pt]
\multirow{2}{*}{Metric} & \multicolumn{2}{c|}{Intersection} & \multicolumn{2}{c}{Non-intersetion} \\ \cline{2-5} 
 & Current SR & Future SR & Current SR & Future SR \\ \hline
Micro Avg Pre & 47.0 & 52.7 & 65.3 & 62.9 \\
Macro Avg Pre & 20.9 & 20.3 & 50.4 & 53.8 \\
mAP & 26.4 & 24.5 & 51.2 & 53.8 \\
\toprule[2pt]
\end{tabular}
\caption{\textbf{Performances of Semantic Region Prediction.} Current SR stands for the current semantic region, while Future SR stands for the future semantic region.}
\label{table:SR}
\end{table}
\label{subsec:exp_srp}

\subsection{Driver Intention Prediction}

\xhdr{Testing Data and Experiment Setup.} After training SRP on the video clips in \cref{subsec:exp_srp}, we further use the intention labels to train the intention classifier. Details are provided in the supplementary materials. We evaluate driver intention prediction models on both HDD~\cite{Ramanishka_behavior_CVPR_2018} test set and nuScenes~\cite{caesar2020nuscenes} datasets. Note that in HDD, there is no overlap between the training data and test data. 
We evaluate models on 1438 sequences in HDD (including 393 interactive scenarios) and 221 sequences in nuScenes. We use the same evaluation metrics as \cref{subsec:exp_srp}.

\xhdr{Baselines and Comparisons.} We implement several baselines with the same image feature extractor as the proposed SRP-INT. LSTM~\cite{hochreiter1997long} is a general-purpose
 sequential modeling methods. OadTR~\cite{wang2021oadtr} takes advantage of the popular Transformers~\cite{dosovitskiy2020image} and is a competitive online/real-time action recognition model.  We also implement LSTM with Exponential Loss (LSTM+EL), as \cite{Jain2016RecurrentNN} shows the effectiveness of Exponential Loss for driver intention prediction. We modify TRN~\cite{XuOnlineICCV2019} to predict trajectories (similar to the work~\cite{evfl}) and use the learned representation for intention prediction. As shown in \cref{table:intention}, we demonstrate favorable performances on both datasets and prove the effectiveness of our framework empirically. Qualitative results are presented in the supplementary materials.

\begin{table*}[t]
\small
\begin{center}

\begin{tabular}{C{0.134\linewidth}|C{0.032\linewidth}|C{0.065\linewidth}|C{0.065\linewidth}|C{0.04\linewidth}|C{0.065\linewidth}|C{0.065\linewidth}|C{0.04\linewidth}|C{0.065\linewidth}|C{0.065\linewidth}|C{0.04\linewidth}}
\toprule[2pt]
 \multirow{2}{*}[-3ex]{Model}  & \multirow{2}{*}[-1.5ex]{Aux} & \multicolumn{3}{c|}{HDD} & \multicolumn{3}{c|}{HDD Interactive} & \multicolumn{3}{c}{nuScenes} \\ \cline{3-11} 
 &  &  
 \begin{tabular}[c]{@{}c@{}}Macro\\Avg Pre\end{tabular} & \begin{tabular}[c]{@{}c@{}}Micro\\Avg Pre\end{tabular} & \begin{tabular}[c]{@{}c@{}}mAP \end{tabular} & \begin{tabular}[c]{@{}c@{}}Macro\\Avg Pre\end{tabular} & \begin{tabular}[c]{@{}c@{}}Micro\\Avg Pre\end{tabular} & \begin{tabular}[c]{@{}c@{}}mAP \end{tabular} & \begin{tabular}[c]{@{}c@{}}Macro\\Avg Pre\end{tabular} & \begin{tabular}[c]{@{}c@{}}Micro\\Avg Pre\end{tabular} & \begin{tabular}[c]{@{}c@{}}mAP \end{tabular} \\ \hline
LSTM\cite{hochreiter1997long} &  -  & 45.0 & 64.9 & 51.5 & 30.8 & 56.2 & 62.4 & 37.3 & \textbf{68.8} & 62.0 \\
LSTM-EL\cite{Jain2016RecurrentNN} &  -  & 45.0 & 65.5 & 52.4 & 29.1 & 51.8 & 60.9 & 35.6 & 62.0 & 61.0 \\
OadTR\cite{wang2021oadtr} &  -  & 35.9 & 24.3 & 36.3 & 48.4 & 46.9 & 54.1 & \textbf{47.8} & 64.3 & 50.7 \\
TRN-Tra &  Tra  & 45.0 & 70.8 & 47.5 & 30.9 & 59.8 & 57.2 & 35.7 & 58.8 & 58.7 \\
SRP-INT &  SR  & \textbf{55.3} & \textbf{73.8} & \textbf{57.9} & \textbf{67.0} & \textbf{70.3} & \textbf{69.5} & 41.1 & 68.3 & \textbf{66.7} \\

\bottomrule[2pt]
\end{tabular}
\caption{\textbf{Quantitative results of driver intention prediction.} We compare SRP-INT with baselines. \textbf{Aux} stands for auxiliary tasks. Tra and SR stand for trajectory and semantic region, respectively. All models have the same feature extractor~\cite{MVD2017}.}
\label{table:intention}
\end{center}
\end{table*}
\begin{table*}[t]
\small
\centering
\begin{tabular}{c|c|c|c|c|c|c}
\toprule[2pt] & \multicolumn{3}{c}{Crossing Vehicle} & \multicolumn{3}{|c}{Crossing Pedestrian} \\ \cline{2-7}
\multirow{-2}{*}{Model} &
  Acc 0.5 & Acc 0.75& mAcc & Acc 0.5 & Acc 0.75 & mAcc \\ \hline
\cite{li2020make} (paper) & 49.2          & 48.6          & 43.0            & 35.7                  & 32.1                  & 27.0                    \\  
\cite{li2020make} (our implementation) &49.2          & 48.2          & 42.7          & 33.3                  & 29.8                  & 26.2                  \\  
SRP-ROI &\textbf{51.8} & \textbf{51.1} & \textbf{45.1} & \textbf{42.9}         & \textbf{39.3}         & \textbf{33.3}   \\

\bottomrule[2pt]
\end{tabular}
\caption{\textbf{Quantitative results of risk object identification.} We evaluate risk object identification models on two risk object categories: Crossing Vehicle and Crossing Pedestrian.} 
\label{table:roi}
\end{table*}
\subsection{Risk Object Identification}

\xhdr{Experimental Setup and Evaluation.} We follow the experiment setup in~\cite{li2020make} and train separate models on two challenging dynamic risk object categories: Crossing Vehicle and Crossing Pedestrian. Like~\cite{li2020make}, we evaluate our models by calculating the IOU between the predicted risk object and ground truth. We report accuracy at IOU thresholds of 0.5, 0.75, and mean accuracy.

\xhdr{Implementation Details.}
 We utilize Mask R-CNN \cite{he2017mask} and DeepSORT \cite{Wojke2017simple} to compute the tracking proposals of risk object candidates. The pre-trained semantic region representation is fused with the ego representation in~\cite{li2020make} after passing through a fully connected layer. In practice, the output dimension of the fully connected layer is 100. In this stage, we train the model using Adam \cite{adam2015} optimizer with default parameters, a learning rate of $0.0001$, and weight decay of $0.0001$. The model is trained for 20 epochs. After training, we follow the inference procedure in \cite{li2020make} to obtain the bounding boxes of the risk object in each frame. We do not apply any heuristic to remove objects from tracking proposals and models are trained separately for each category.

\xhdr{Quantitative Results.}
We compare our method with ~\cite{li2020make}. The quantitative results show that our model obtain favorable performance compared to~\cite{li2020make}, which demonstrates that semantic region prediction can help risk object identification. Qualitative results are presented in the supplementary materials.


\section{Limitations}
Although we have shown the effectiveness of our proposed representation, some limitations need further exploration. First, our proposed semantic regions cannot be applied to complicated topologies like roundabouts or other real-world edge cases in intersections.  Possible solutions are: defining the semantic regions of all intersections by the number of branches of the intersections and considering one roundabout as a series of 3-way intersections.
Second, learning semantic regions from egocentric view images alone is challenging.
Additionally, the performance of semantic region prediction at intersections is unsatisfactory. 
To improve the performance, we could consider incorporating Bird-Eye-View representation~\cite{bev_icra2022}. Third, we have not truly associated images with semantic regions. Instead of predicting the label of semantic regions, we could consider an encoder-decoder based model to predict the current/future scene representations~\cite{epc}.


\section{Conclusion}

In this work, we study the problem of road scene-level representation learning from egocentric videos for driver intention prediction and risk object identification. 
We propose a novel representation called semantic region, which aims to capture higher-level semantic and geometric representations of traffic scenes around ego vehicles while performing actions to their destination.
We cast representation learning as semantic region prediction and
propose an automatic semantic region labeling algorithm for egocentric videos collected in intersections.
%
We demonstrate the effectiveness of the learned representation on real-world datasets, i.e., HDD and nuScenes.
In particular, the learned representation can generalize to unseen data (i.e., nuScenes dataset) without finetuning the driver intention prediction task.
%
%
%
We hope that our findings will pave the way for further advances in road scene-level representation learning from egocentric views for downstream tasks such as planning and decision-making. 
%
%
\label{sec:conclusion}

\acknowledgments{
A part of the work was done when Z. Xiao was an intern and Y.-T. Chen
was a research scientist at Honda Research Institute USA, San Jose, CA, USA. The work is partly sponsored by Honda Research Institute USA. Yi-Ting Chen is supported in part by the Higher Education Sprout Project of the National Yang Ming Chiao Tung University and Ministry of Education (MOE), the Ministry of Science and Technology (MOST) under grants 110-2222-E-A49-001-MY3 and 110-2634-F-002-051, and Mobile Drive Technology Co., Ltd (MobileDrive).}



\bibliography{example}  

\newpage
\beginsupplement
\section{Automatic Semantic Region Labeling}
%

\begin{figure}[h]
\centering
\includegraphics[width = 0.7\columnwidth]{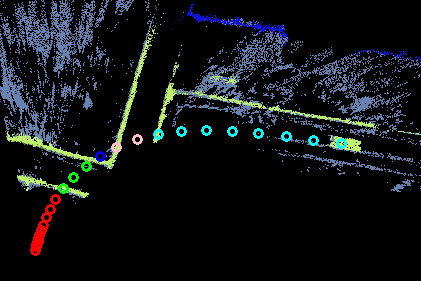}
\caption{\textbf{Sample of automatic labeling.} We show the results of automatic semantic region labeling of a \textit{Right-turn} sample. The semantic regions are visualized in colors. The {\color{red} red} circle indicates $S$. The {\color{green} green} circle indicates $A_3$. The {\color{blue} blue} circle indicates $B_3$. The {\color{pink} pink} circle indicates $C_3$. The {\color{cyan} cyan} circle indicates $T_3$. Best Viewed in color.}
\label{fig:semantic_label}
\end{figure}

%
%
We use a \textit{Right-turn} sample to illustrate the automatic semantic region labeling process. 
We derive a semantic BEV image with the process described in the main paper.
As shown in Figure~\ref{fig:semantic_label},  camera locations overlapping with the first and second crosswalks are annotated as $A_i$ and $C_i$, respectively. 
The poses locating between $A_{i}$ and $C_{i}$ are annotated as $B_{i}$.
Camera poses locating in areas before the first and the second \textit{crosswalk} are $S_{i}$ and $T_{i}$, respectively. 
Note that the parameter $i$ is 3 because this is a \textit{Right-turn} at a 4-way intersection.
The parameter $i$ is set to 1 and 2 for \textit{Left-turn} and \textit{Go Straight}, respectively.

%
%
%

To evaluate the effectiveness of our automatic semantic region labeling process, we randomly pick 100 video clips and annotate ground truth semantic regions manually. 
The accuracy of automatic semantic labeling is 76.4\%.
We diagnose the results and find the following reasons for failures.
First, some video clips do not start from the semantic region $S$ because the original starting time labeled in the HDD dataset is inaccurate.
Second, lines like lane-changing lines and arrows indicating directions are wrongly predicted as crosswalks by the segmentation model. 
To improve the quality of labeling, we plan to annotate the center of the 4-way intersection (i.e., $B_i$) and train another semantic segmentator to mitigate the second issue in future work.
%

\section{Experimental Details}
\xhdr{Driver Intention Prediction.}
After training SRP, we freeze every other layers but the intention classifier. Similar to training SRP, We use Adam optimizer \cite{adam2015} with default parameters, a learning rate of $0.0001$, and weight decay of $0.0005$. The model is trained for 60 epochs. We report the performances of the last epoch.

\xhdr{Risk Object Identification.} We make use of the same weights of SRR as SRP-INT does and the weights are frozen during the training process. The hidden state of the SRP is connected to a fully connected(FC) layer before being fussed with the ego representation. We then follow the two-stage strategy as described in the main paper to obtain the risk object predictions.

\section{Ablation Study: Model Pretraining}

We evaluate the impact of model pretraining for the base model. We follow the same training procedure as the main paper and diagnose our SRP-INT on the HDD, HDD interactive and nuScenes.

As shown in \cref{table:abl_pf} and \cref{table:abl_pf_2}, the backbone model pretrained on the Mapillary Vistas dataset results in significantly better performance compared with the backbone models trained on other datasets. Note that these backbones are trained on different tasks. The Mapillary Vistas backbone is pre-trained on the panoptic segmentation task. The nuScenes backbone is trained on instance segmentation task using data released in nuImage, an extension of nuScenes that contains additional images and 2D annotations. Note that they have semantic labels for the drivable surface. The COCO Panoptic backbone is trained on COCO Panoptic Segmentation. The model performs favorably in different settings, while COCO Panoptic is not a traffic scene dataset. The tables show that the in-domain nuScenes backbone cannot perform well in many metrics. We hypothesize that the two tasks, i.e., intention prediction and risk object identification, require \textit{Stuff} information (e.g., road, lane marking, and crosswalk). On the other hand, the nuScenes backbone learns to detect objects, which could explain the superior performance on HDD interactive cases because these cases involve interaction with other traffic participants.

\begin{table*}[t]
\small
\begin{center}

\begin{tabular}{C{0.09\linewidth}|C{0.15\linewidth}|C{0.065\linewidth}|C{0.065\linewidth}|C{0.04\linewidth}|C{0.065\linewidth}|C{0.065\linewidth}|C{0.04\linewidth}}
\toprule[2pt]
 \multirow{2}{*}[-3ex]{Model}  & \multirow{2}{*}[-3ex]{Feature} & \multicolumn{3}{c|}{HDD} & \multicolumn{3}{c}{HDD Interactive} \\ \cline{3-8} 
 &  &  
 \begin{tabular}[c]{@{}c@{}}Macro\\Avg Pre\end{tabular} & \begin{tabular}[c]{@{}c@{}}Micro\\Avg Pre\end{tabular} & \begin{tabular}[c]{@{}c@{}}mAP \end{tabular} & \begin{tabular}[c]{@{}c@{}}Macro\\Avg Pre\end{tabular} & \begin{tabular}[c]{@{}c@{}}Micro\\Avg Pre\end{tabular} & \begin{tabular}[c]{@{}c@{}}mAP \end{tabular} \\ \hline
SRP-INT &  ImageNet  & 51.3 & \textbf{74.6} & 53.9 & 45.3 & 59.3 & 60.1 \\
SRP-INT &  nuScenes  & 25.0	& 57.8 & 45.5 & \textbf{67.4} & 69.8	& 69.1 \\
SRP-INT &  COCO Panoptic  & 52.8 & 61.8	& 51.7 & 46.1 & 61.1 & 63.3 \\
SRP-INT &  Mapillary Vistas & \textbf{55.3} & 73.8 & \textbf{57.9} & 67.0 & \textbf{70.3} & \textbf{69.5} \\

\bottomrule[2pt]
\end{tabular}
\caption{\textbf{Ablation study for model pretraining in HDD dataset on driver intention prediction.} Base model pretrained on the Mapillary Vistas dataset leads to better performance in general. The results confirm the importance of the final task of a pretraining model.}
\label{table:abl_pf}
\end{center}
\end{table*}

\begin{table*}[t]
\small
\begin{center}

\begin{tabular}{C{0.09\linewidth}|C{0.15\linewidth}|C{0.065\linewidth}|C{0.065\linewidth}|C{0.04\linewidth}}
\toprule[2pt]
 \multirow{2}{*}[-3ex]{Model}  & \multirow{2}{*}[-3ex]{Feature} & \multicolumn{3}{c}{nuScenes}  \\ \cline{3-5} 
 &  &  
 \begin{tabular}[c]{@{}c@{}}Macro\\Avg Pre\end{tabular} & \begin{tabular}[c]{@{}c@{}}Micro\\Avg Pre\end{tabular} & \begin{tabular}[c]{@{}c@{}}mAP \end{tabular}  \\ \hline
SRP-INT &  ImageNet & 36.0 & 59.7 & 58.1  \\
SRP-INT &  nuScenes & \textbf{45.1} & 37.6 & 59.5 \\
SRP-INT &  COCO Panoptic &  37.6 & 63.8 & 61.1  \\
SRP-INT &  Mapillary Vistas&  41.1 & \textbf{68.3} & \textbf{66.7}  \\

\bottomrule[2pt]
\end{tabular}
\caption{\textbf{Ablation study for model pretraining in nuScenes dataset on driver intention prediction.} Base model pretrained on the Mapillary Vistas dataset leads to better performance in general. The results confirm the importance of the final task of a pretraining model.}
\label{table:abl_pf_2}
\end{center}
\end{table*}




\section{Generalization to nuScenes}
\begin{figure}[h]
     \centering
     \begin{subfigure}[b]{0.6\textwidth}
         \centering
         \includegraphics[width=\textwidth]{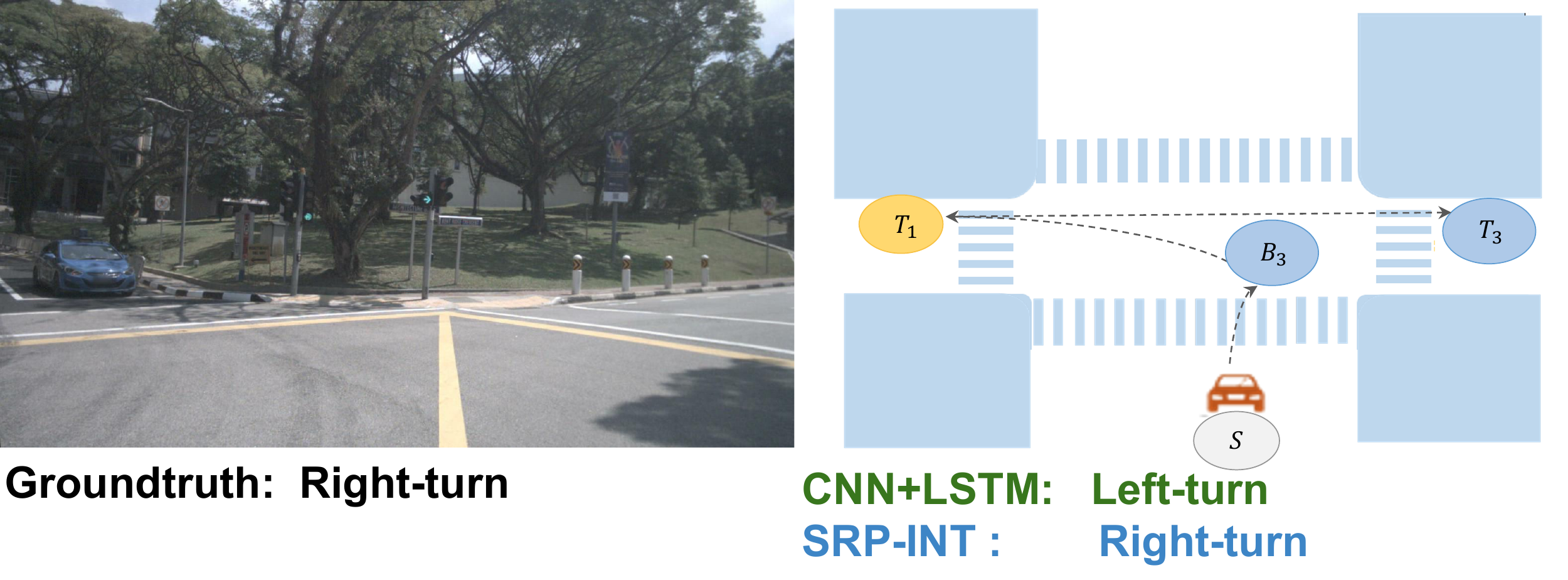}
         \caption{Left-hand Traffic}
         \label{fig:failure1}
     \end{subfigure}
     \hfill
     \begin{subfigure}[b]{0.6\textwidth}
         \centering
         \includegraphics[width=\textwidth]{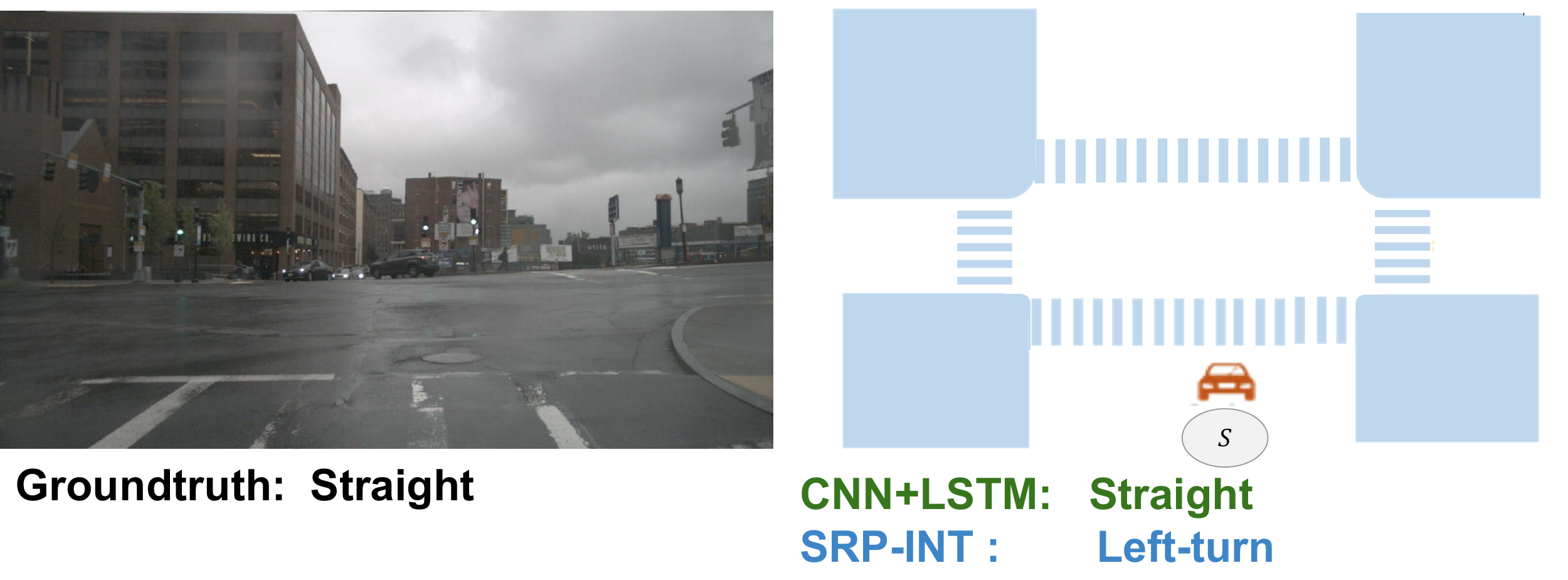}
         \caption{Empty Intersection}
         \label{fig:failure2}
     \end{subfigure}
        \caption{\textbf{Failure cases on the nuScenes dataset.} We show two typical failure cases of SRP-INT on the nuScenes dataset. We provide ground truth as well as the driver intention predictions of the CNN+LSTM baseline and our proposed SRP-INT. The semantic region predictions are shown on the right side.}
        \label{fig:differentcases}
\end{figure}

In the task of driver intention prediction, to demonstrate the effectiveness of the learned representations, we first train our model on the HDD dataset \cite{Ramanishka_behavior_CVPR_2018} and test on the nuScenes dataset~\cite{caesar2020nuscenes} without finetuning.
It is worth noting that the domain gap between the HDD dataset and the nuScenes dataset is significant and could lead to false predictions in either semantic region predictions or intention predictions.
%
Some videos in the nuScenes dataset are collected in countries with left-hand traffic, while data in HDD dataset is in right-hand traffic conditions.
%
Another typical failure occurs when the ego vehicle approaches an empty 4-way intersection. 
As shown in \cref{fig:failure2}, it is challenging to make correct predictions without additional cues.
One possible future direction is to leverage drivers’ gazes as in the Brain4Car project~\cite{Jain_driver_maneuver_iccv2015} or steering signals. 

\section{Interactive Scenarios}
It is challenging to predict the driver's intention in the modality of monocular image sequences due to complicated driving scenarios such as drivers may have to stop for crossing vehicles or yield to crossing pedestrians before they reach their intended goals. We call these cases interactive scenarios. We evaluate our model on interactive scenarios on the HDD testing set because of their importance in real-world applications.
We present quantitative and qualitative evaluations of interactive scenarios in the main paper as well as the following sections of the supplementary materials.

\section{Qualitative Results}
\begin{figure*}[t]
\centering
\includegraphics[width = 0.82\linewidth]{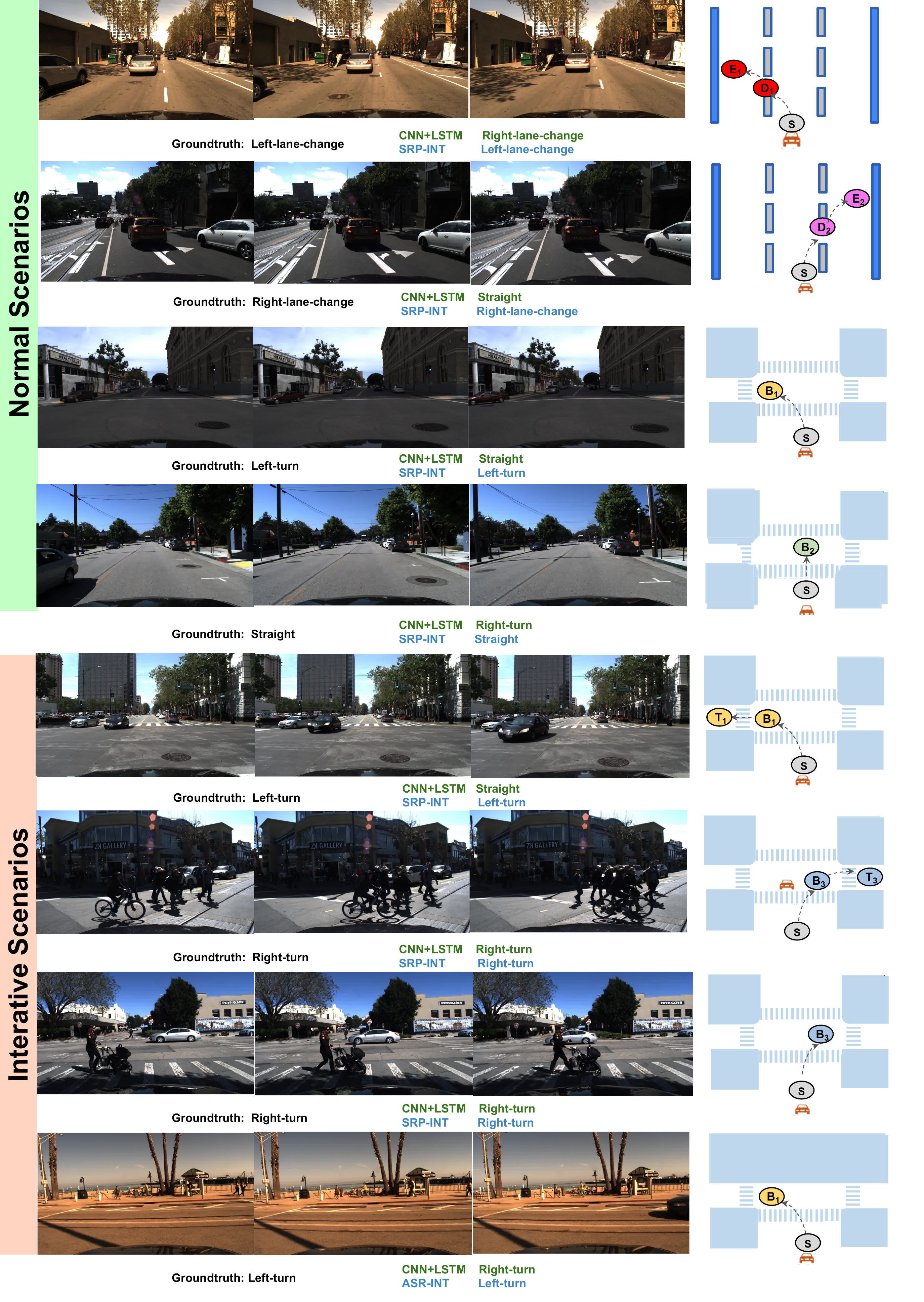}
\caption{\textbf{Qualitative results of driver intention prediction on the HDD dataset~\cite{Ramanishka_behavior_CVPR_2018}.} The examples shown in the first four rows indicate normal cases, i.e., the ego-vehicle navigating through the intersection without interactions with other traffic participants. 
The examples shown in the last four rows are interactive scenarios. 
We provide the ground truth of ego-vehicle intention and the prediction of the final SRP-INT and the CNN+LSTM baseline on the HDD dataset. 
The predictions of semantic regions of SRP-INT are displayed on the right side of each scenario, where traffic participants intervene in the movement of the ego-vehicle.
The results demonstrate the proposed framework can predict semantic regions reliably, and that helps the visual system predict ego-vehicle intention.
The qualitative experiments empirically justify the value of the proposed scene-level representation learning.}
\label{fig:QualiIntention_HDD}
\end{figure*}

\begin{figure*}[t]
\centering
\includegraphics[width = 0.8\linewidth]{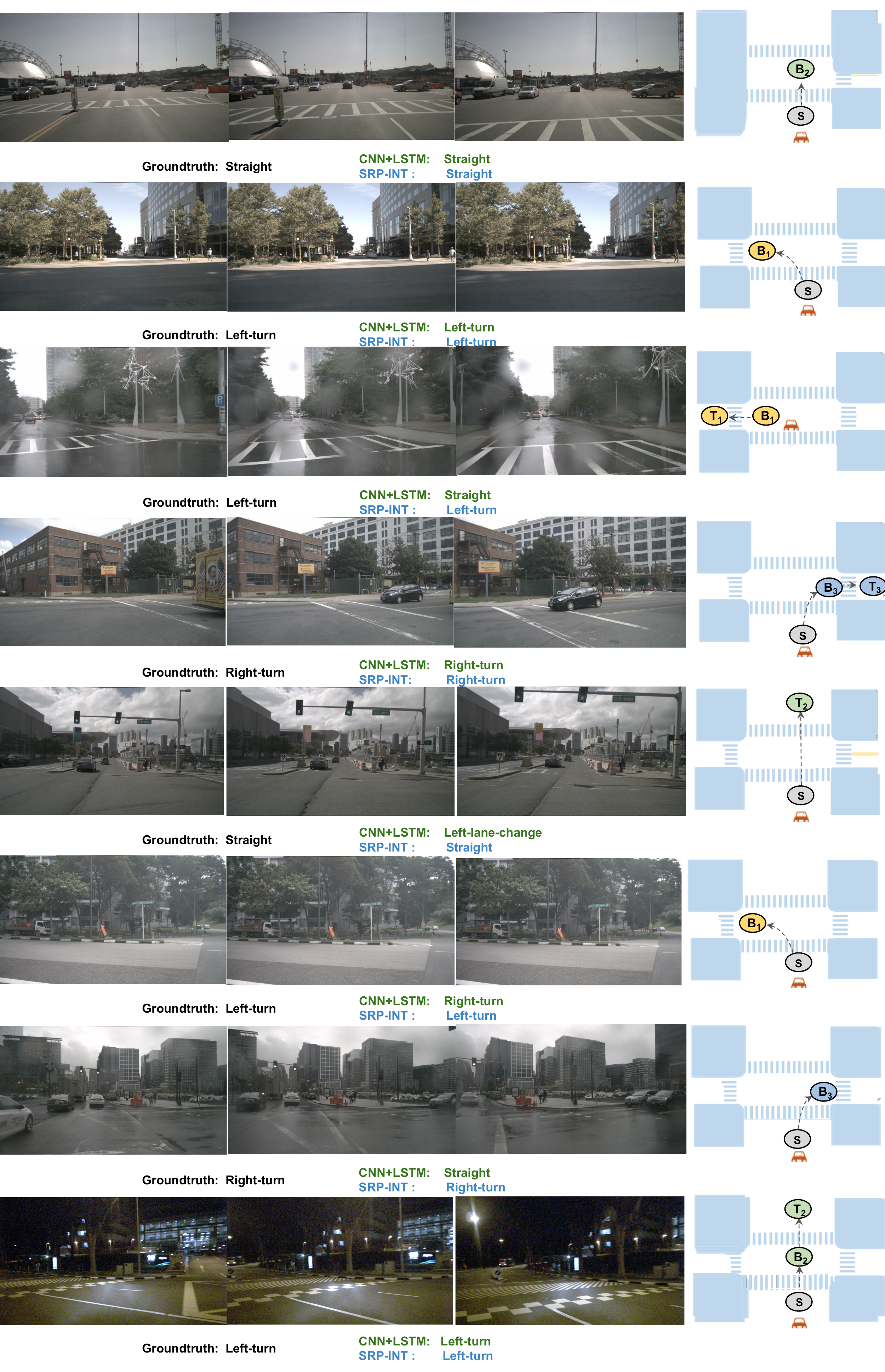}
\caption{\textbf{Qualitative results of driver intention prediction on the nuScenes dataset~\cite{caesar2020nuscenes}.} 
Similar to the results on the HDD dataset~\cite{Ramanishka_behavior_CVPR_2018} shown in Figure~\ref{fig:QualiIntention_HDD}, the ground truth of ego-vehicle intention, as well as the prediction of the SRP-INT and the CNN+LSTM baseline on the nuScenes dataset, are presented.
The semantic region predictions are provided on the right side of each case.}
\label{fig:QualiIntention_nuScenes}
\end{figure*}

We show qualitative results of driver intention prediction on the HDD dataset and nuScenes dataset of SRP-INT in \cref{fig:QualiIntention_HDD} and \cref{fig:QualiIntention_nuScenes}, respectively. For comparison, we also show the intention predictions of the CNN+LSTM baselines as the intention ground truth.

\begin{figure*}[t]
\centering
\includegraphics[width = \linewidth]{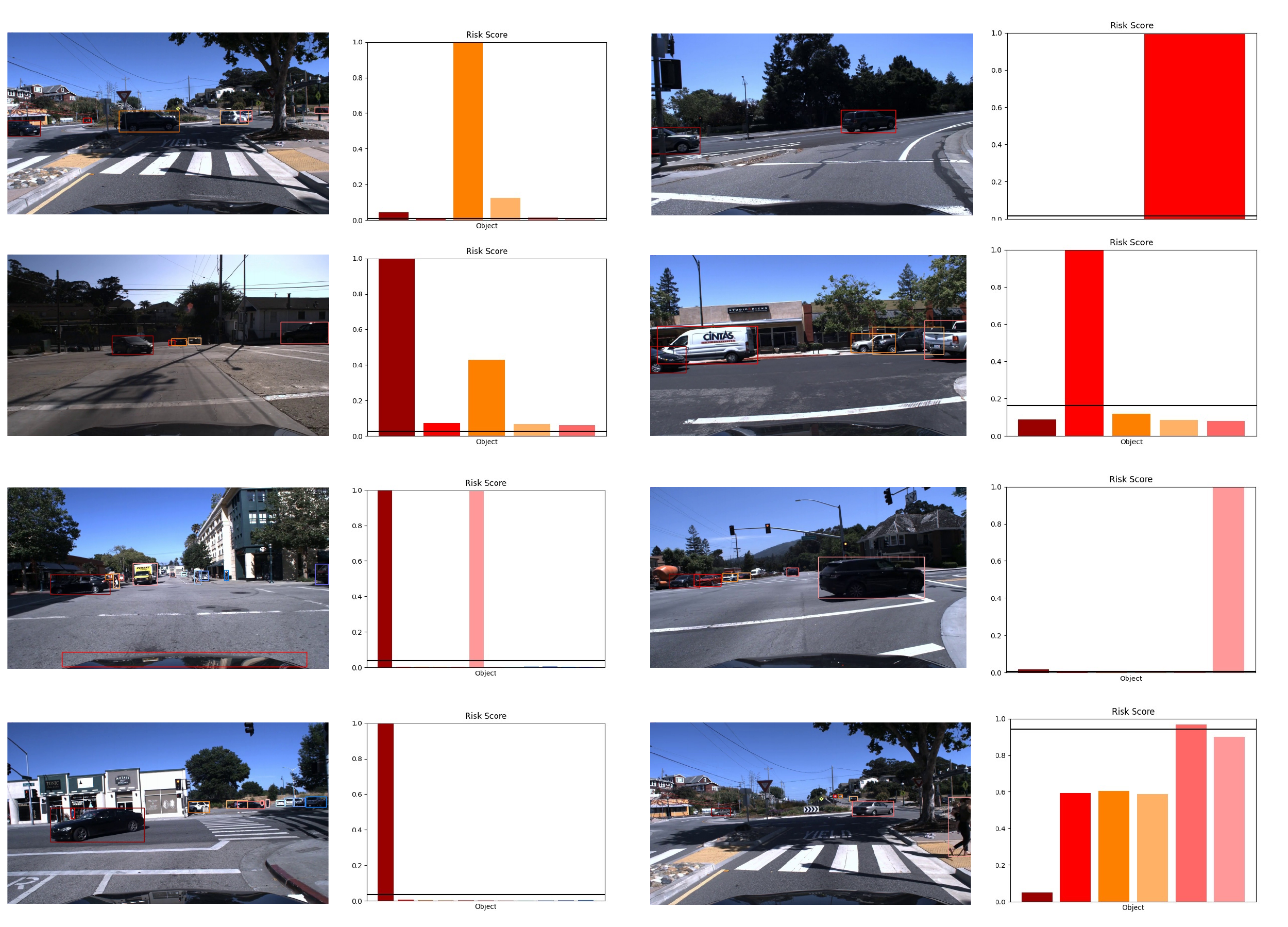}
\caption{\textbf{Qualitative results of risk object identification -- Crossing-Vehicle.} We demonstrate the effectiveness of the proposed scene-level representation for risk object identification. According to the definition of the risky object proposed in~\cite{li2020make}, the candidate with the highest risk score is the risk object. 
%
In this figure, we show the risk scores of each object candidate and demonstrate the system can differentiate risk and non-risk objects in various crossing vehicle scenarios. For each candidate, the color of the bar matches the color of the bounding box.}
\label{fig:Quali_ROI_vehicle}
\end{figure*}

\begin{figure*}[t]
\centering
\includegraphics[width = \linewidth]{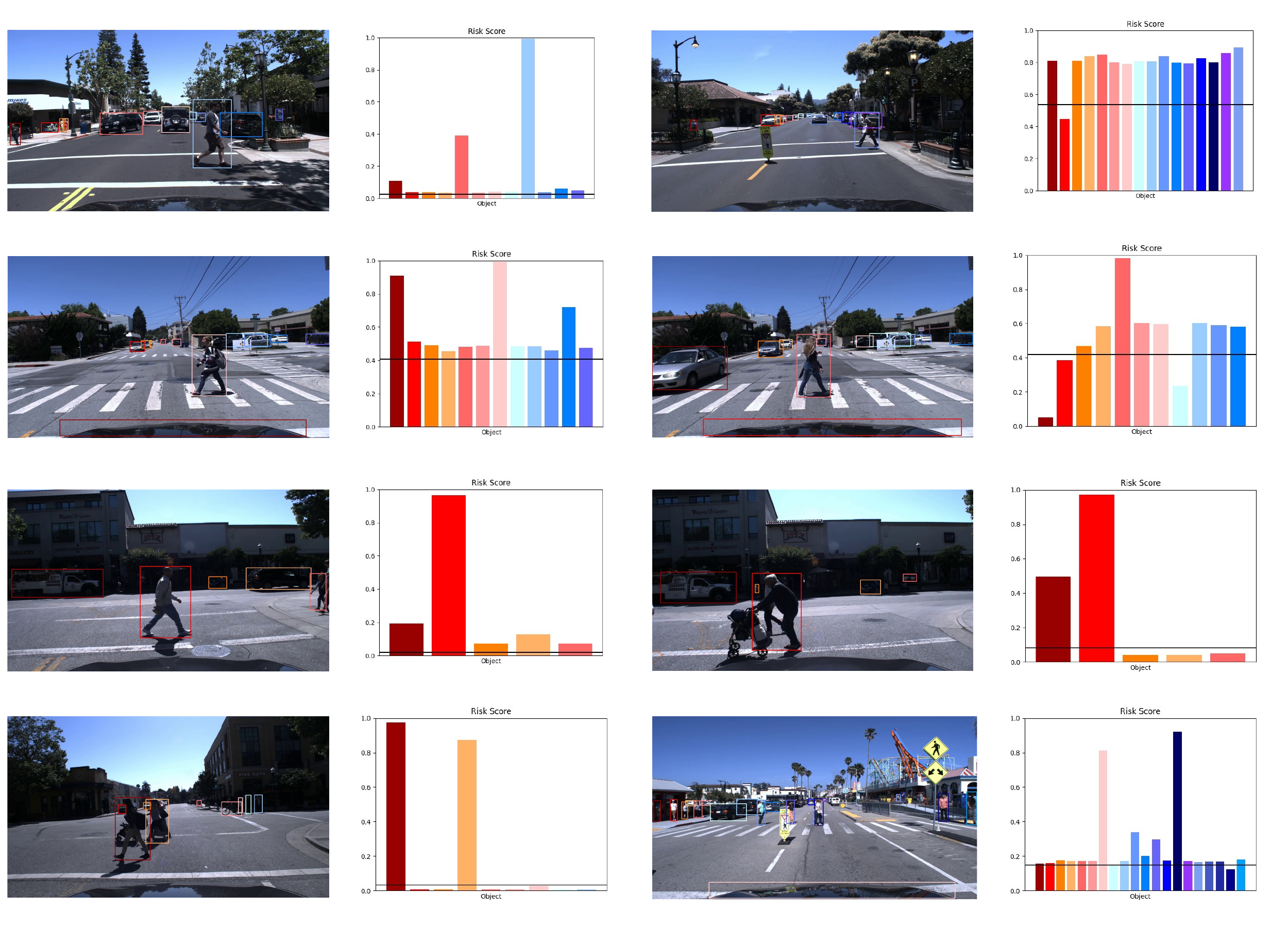}
\caption{\textbf{Qualitative results of risk object identification -- Crossing-Pedestrian.} We demonstrate the effectiveness of the proposed scene-level representation for risk object identification. 
According to the definition of the risky object proposed in~\cite{li2020make}, the candidate with the highest risk score is the risk object. 
%
%
In this figure, we show the risk scores of each object candidate and demonstrate the system can differentiate risk and non-risk objects in various crossing pedestrian scenarios. 
For each candidate, the color of the bar matches the color of the bounding box.}
\label{fig:Quali_ROI_pedestrian}
\end{figure*}

Qualitative results of SRP-ROI on two risk object categories: Crossing Vehicle and Crossing Pedestrian are shown in \cref{fig:Quali_ROI_vehicle} and \cref{fig:Quali_ROI_pedestrian}, respectively. 

\end{document}